\documentclass[10pt,twocolumn,letterpaper]{article}

\usepackage[pagenumbers]{cvpr} 

\definecolor{cvprblue}{rgb}{0.21,0.49,0.74}
\usepackage[pagebackref,breaklinks,colorlinks,allcolors=cvprblue]{hyperref}

\def\paperID{8} 
\def\confName{CVPR -  CVEU}
\def\confYear{2026}

\usepackage{xspace}
\usepackage{multirow}
\usepackage{multicol}
\usepackage{diagbox}
\usepackage{pifont} 
\usepackage{verbatim} 

\usepackage{listings}
\usepackage{tcolorbox}
\usepackage{tocloft}

\newcommand{\ourmodel}{\textbf{\texttt{LaDe}}\xspace}

\newcommand{\design}{media design\xspace}
\newcommand{\designs}{media designs\xspace}
\newcommand{\Design}{Media design\xspace}
\newcommand{\Designs}{Media designs\xspace}

\title{\ourmodel: Unified Multi-Layered Graphic Media Generation and Decomposition}

\author{Vlad-Constantin Lungu-Stan,  
Ionuţ Mironică,   
Mariana-Iuliana Georgescu \\
Adobe Research, Romania\\
{\tt\small \{vlungustan,mironica,mgeorgescu\}@adobe.com}
}

\begin{document}

\twocolumn[{
 \renewcommand\twocolumn[1][]{#1}
 \maketitle
 \begin{center}
    \centering
    \captionsetup{type=figure}
    \includegraphics[width=\textwidth]{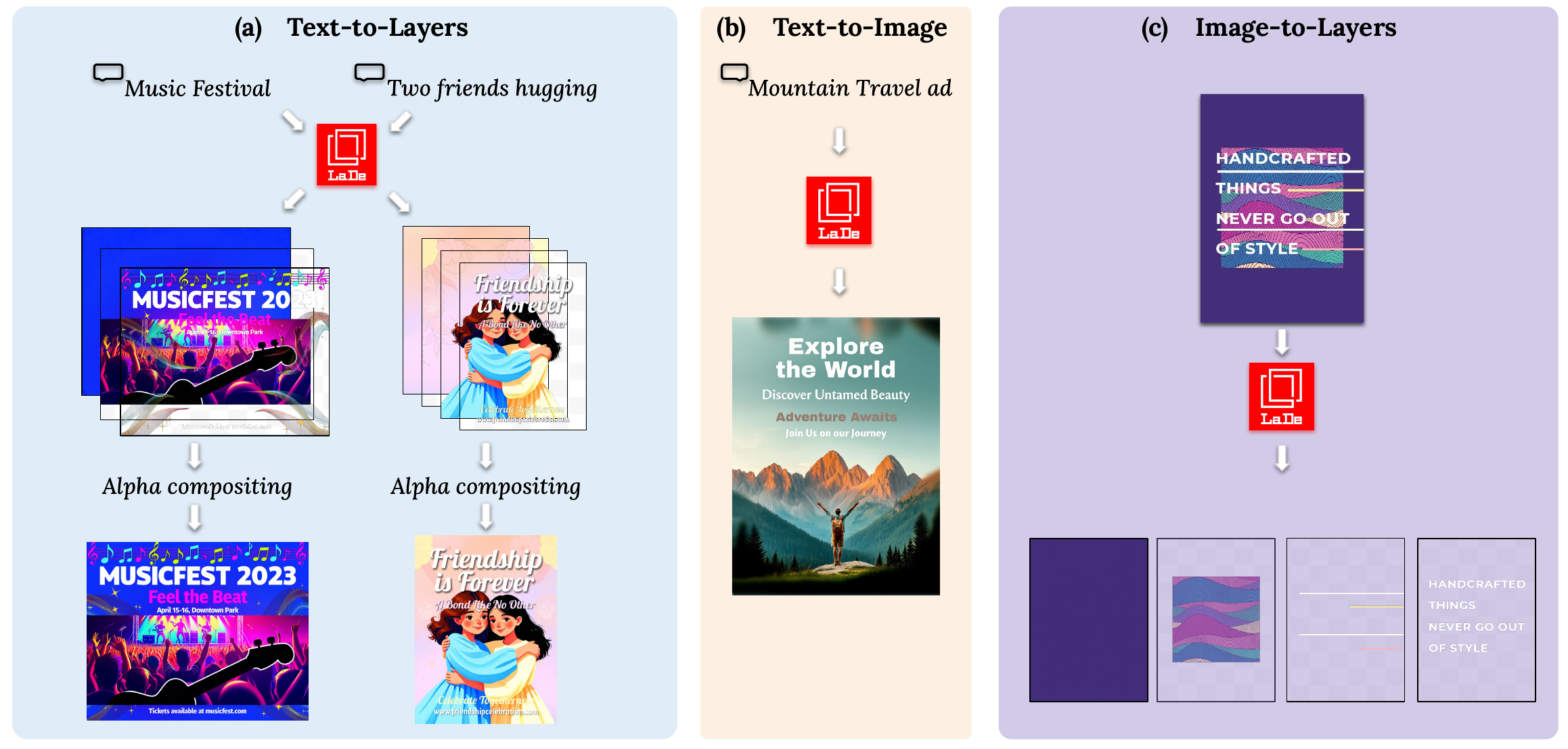}
    \captionof{figure}{The unified \ourmodel performs (a) Text-to-Layers generation in the RGBA space and (b) Text-to-Image generation based on a prompt, (c) Image-to-Layers generation  in the RGBA space given a image. \ourmodel works with variable aspect ratio and number of layers.
     }
    \label{fig:teaser}
 \end{center}
 \vspace{0.1in}
}]

\begin{abstract}
\Design layer generation enables the creation of fully editable, layered design documents such as posters, flyers, and logos using only natural language prompts. Existing methods either restrict outputs to a fixed number of layers or require each layer to contain only spatially continuous regions, causing the layer count to scale linearly with design complexity. We propose \ourmodel (\textbf{La}yered Media \textbf{De}sign), a latent diffusion framework that generates a flexible number of semantically meaningful layers. \ourmodel combines three components: an LLM-based prompt expander that transforms a short user intent into structured per-layer descriptions that guide the generation, a Latent Diffusion Transformer with a 4D RoPE positional encoding mechanism that jointly generates the full \design and its constituent RGBA layers, and an RGBA VAE that decodes each layer with full alpha-channel support. By conditioning on layer samples during training, our unified framework supports three tasks: text-to-image generation, text-to-layers \design generation, and \design decomposition. We compare \ourmodel to Qwen-Image-Layered on text-to-layers and image-to-layers tasks on the Crello test set. \ourmodel outperforms  Qwen-Image-Layered in text-to-layers generation by improving text-to-layer alignment, as validated by two VLM-as-a-judge evaluators (GPT-4o mini and Qwen3-VL).


\end{abstract}    
\section{Introduction}

\begin{figure*}
      \centering
    \includegraphics[width=0.9\linewidth]{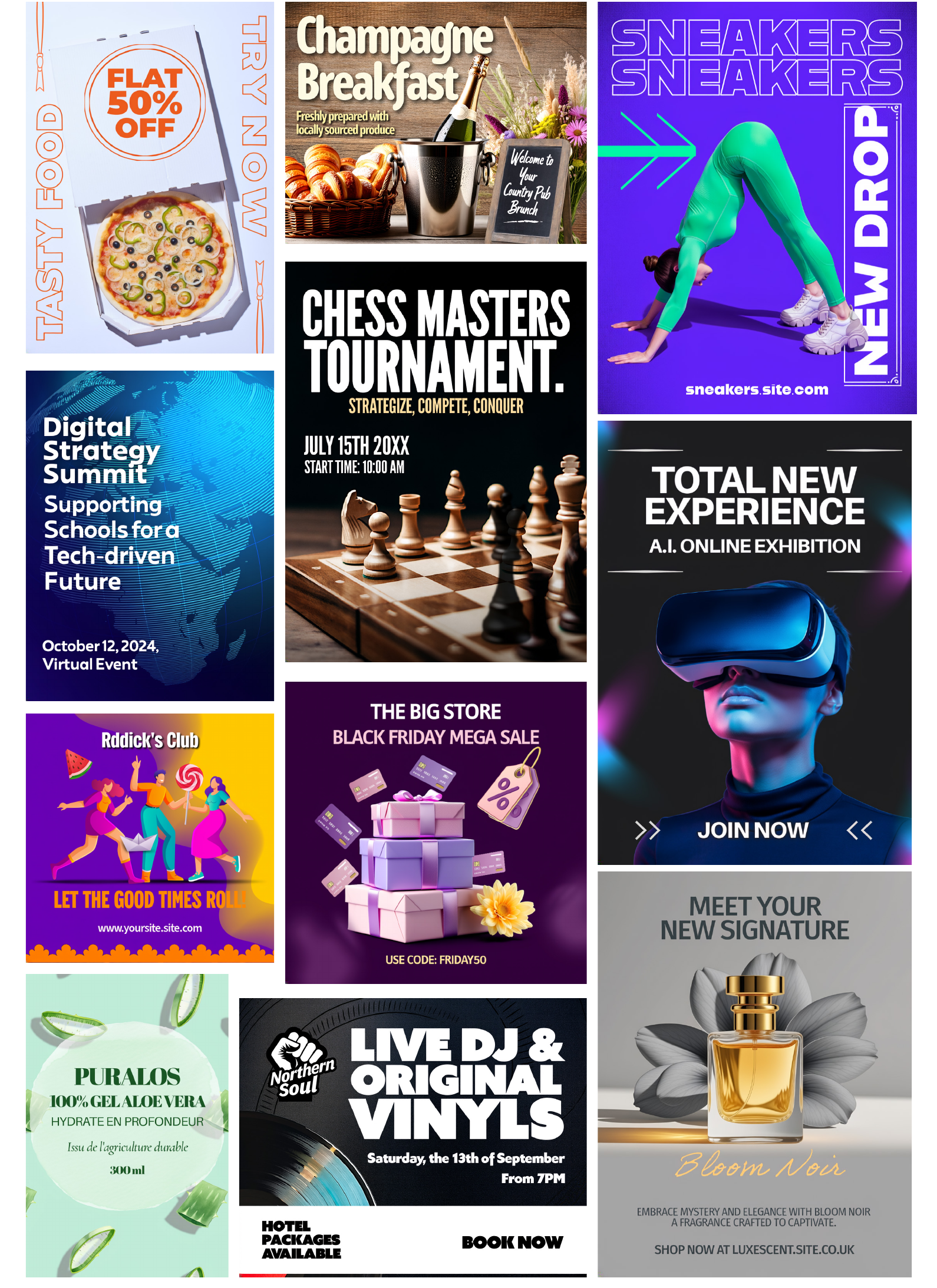}
    \caption{Text-to-Image Generation with \ourmodel.}
    \label{fig:t2i}
\end{figure*}

\begin{figure*}
      \centering
    \includegraphics[width=0.9\linewidth]{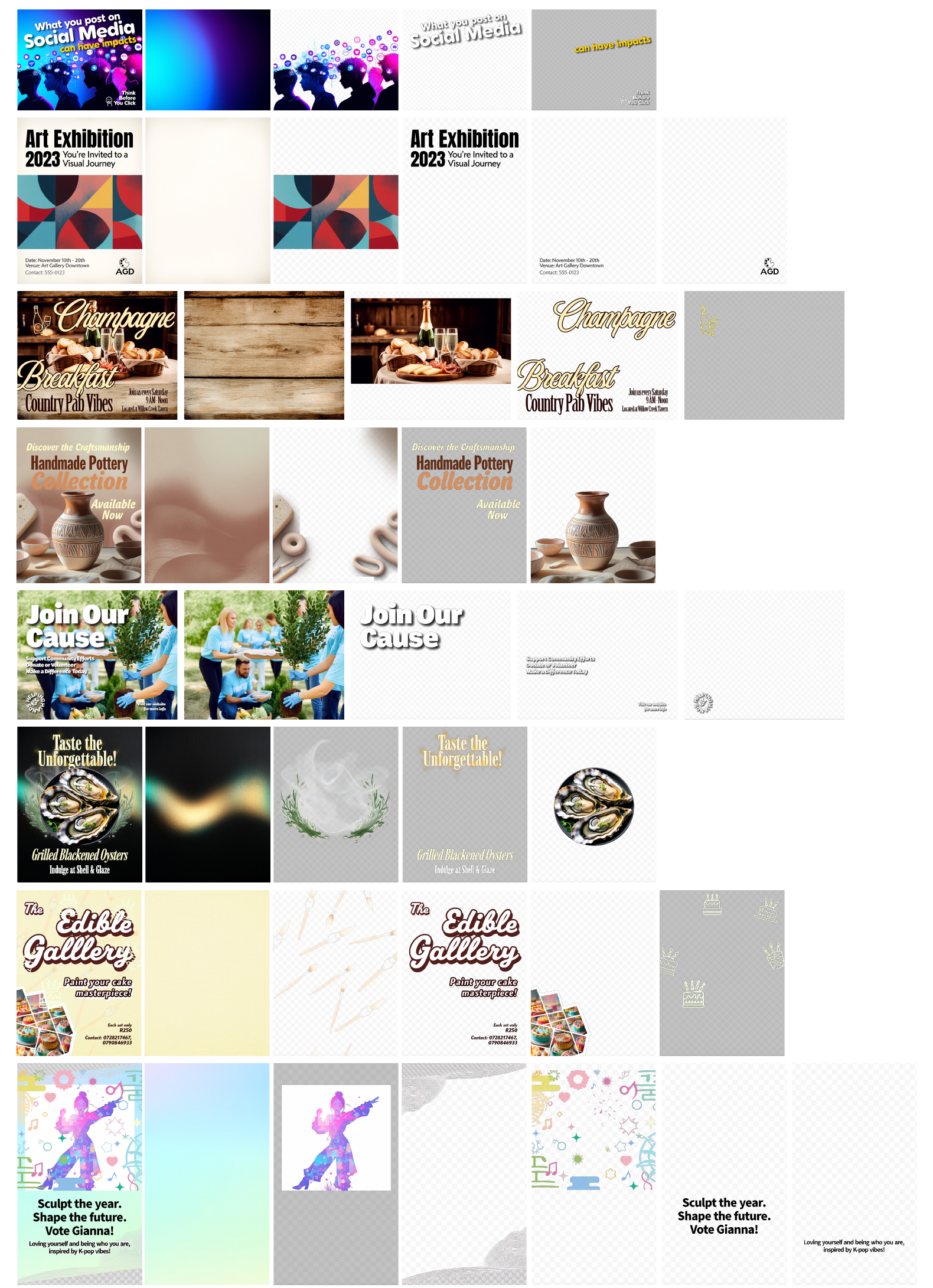}
    \caption{Text-to-Layers Generation with \ourmodel. The gray background is added to emphasize the while content.}
    \label{fig:t2l}
\end{figure*}
 
\begin{figure*}
   \centering
    \includegraphics[width=0.99\linewidth]{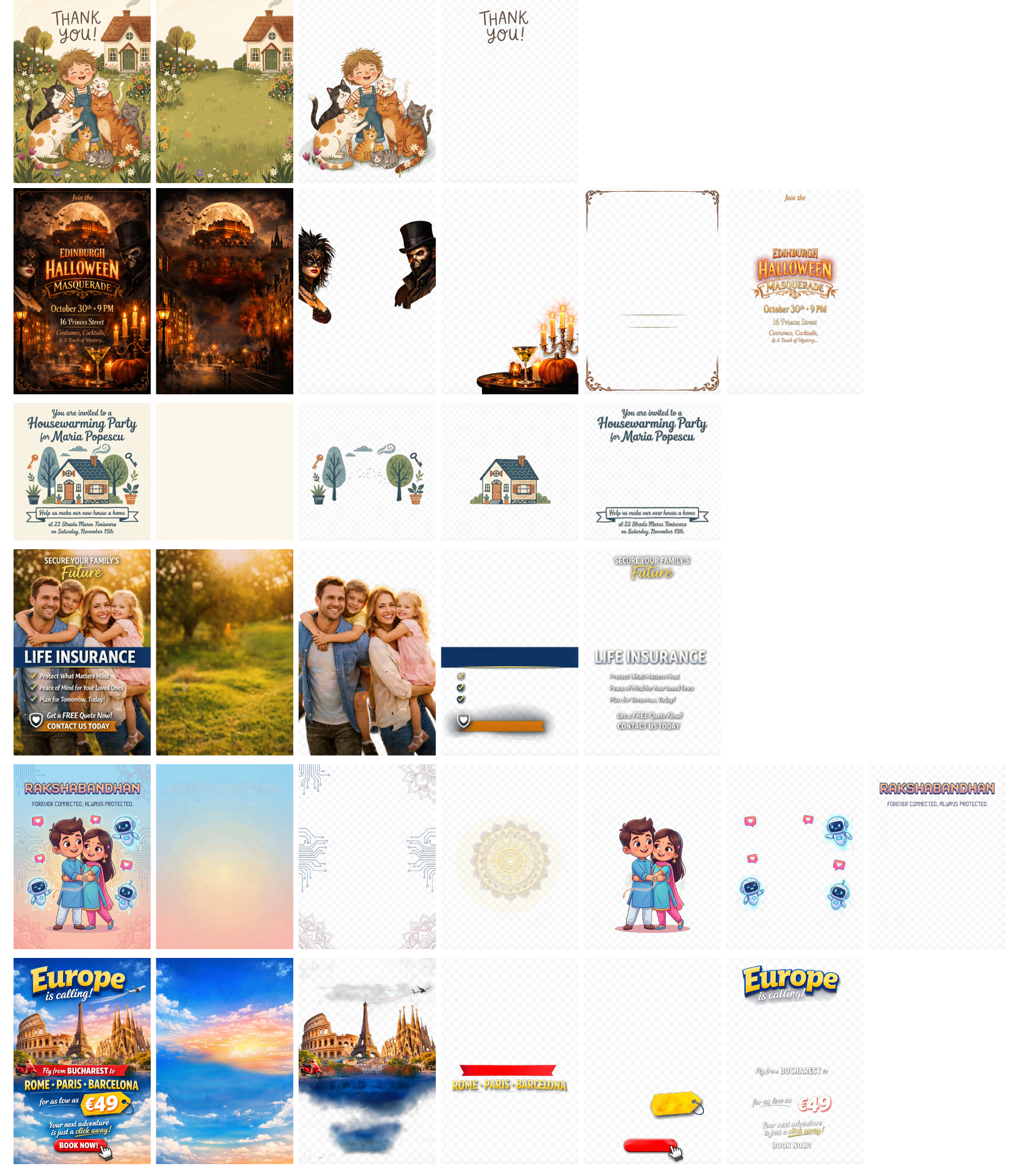}
    \caption{Image-to-Layers Decomposition with \ourmodel. The samples are generated by two state-of-the-art proprietary GenAI frameworks called through their APIs. \\ \\ \\}  
    \label{fig:i2l}
\end{figure*}

The generation of content through generative models has received massive attention recently~\cite{art,yin2025qwenimagelayered}. Diffusion Models~\cite{ho2020denoising, Rombach-CVPR-2022, esser2024scaling, 10657078} (DM) have enabled the creation of images and videos that were not possible with the previous generation of generative models, namely Generative Adversarial Networks~\cite{karras2020analyzing}. Most generative design systems~\cite{Wang2025DesignDiffusionHT,inoue2024opencole,jia2024colehierarchicalgenerationframework} treat a design as a single flat image artifact. However, professional design practice has always been layered and compositional. A poster, advertisement, app screen, or marketing banner is not a monolithic image, it is, in fact, a stack of semantically distinct elements (e.g.: background imagery, graphic shapes, typography, logos, and overlays), each independently editable, replaceable, and purposeful. Although diffusion models have demonstrated remarkable capability in image synthesis, they generate the entire image in a single pass, offering limited control over individual elements. In professional design workflows, this control is exercised through layers that are discrete, blendable components that together compose the final \design. Therefore, we focus on the problem of design layer generation: building a system capable of generating designs not as flat images but as structured, layered compositions that reflect how designs are actually created and used.

Image editing approaches~\cite{art, yin2025qwenimagelayered, liu2025omnipsd} built on diffusion models have attempted to address this, but still lack the fine-grained control that \design generation demands.  Anonymous Region Transformer (ART)~\cite{art} is one of the first frameworks to achieve design layer generation. ART used finetuned Large Language Model (LLM) as a planner to generate bounding boxes that are treated as layers based on the prompt. On the other hand, OmniPSD~\cite{liu2025omnipsd} generates the layers together with the entire (alpha-blended) \design end-to-end without external information.

In fact, existing methods have driven the creation of \design from a single image to several individual layers, increasing the control of \design generation. However, professional designers require a flexible number of layers that do not necessarily scale linearly with the complexity of the design. For example, OmniPSD~\cite{liu2025omnipsd} generates only three layers without the possibility of increasing or decreasing this fixed number. On the opposite side,  ART~\cite{art} is capable of generating up to 50 layers; however, it has the constraint that a layer contains only continuous regions, meaning that an image with 30 small scattered stars would be composed of 30 layers. Therefore, complex designs have a large number of layers, since it splits similar patterns (e.g., the stars, or confetti) into different layers, making the \designs harder to edit while also losing their visual hierarchy. 

We propose \textbf{La}yered Media \textbf{De}signs (\ourmodel) that is capable of generating a flexible number of layers, without increasing the number of layers with the complexity of \design. 
Our system has three main components, namely the prompt expander, the diffusion model, and the RGBA Variational Autoencoder (VAE). \ourmodel requires only a brief prompt that describes the user's intent. The Prompt Expander creates content information using a language known by the diffusion model (also used during training). The diffusion model includes a 4D RoPE~\cite{Su-NC-2024} positional encoding that links the content information to its respective layer. To create a \design the DM takes the encoded content information and noise and generates the full \design together with its constituent RGBA layers. Finally, the VAE model decodes the layers and the full-design one by one in the RGBA space. 
To enable the generation of flexible number of layers while also efficiently utilizing the GPU memory, we propose the \texttt{bucketing} and \texttt{packing} operations that group together samples of similar size.

\ourmodel is trained on a dataset containing layered \designs. To condition the diffusion model on the text prompt, we employ a captioning model to generate textual descriptions for each \design and layers, respectively. Since \ourmodel is trained end-to-end,  our unified single model is capable of performing text-to-layers and text-to-image generation, and image-to-layers decomposition, as illustrated in Figure~\ref{fig:teaser}.

We perform multiple experiments on the Crello test subset~\cite{Yamaguchi-ICCV-2021} that contains 500 user prompts along with their \designs. We compare our framework with Qwen-Image-Layered~\cite{yin2025qwenimagelayered} on the text-to-layers generation and image-to-layers decomposition tasks. We report PSNR (Peak Signal-to-Noise Ratio), RGB L1 and VLM-as-a-judge scores for image-to-layers decomposition, while reporting results using VLM-as-a-judge with two state-of-the-art  Vision Language Models (VLMs) (GPT-4o mini~\cite{openai2024gpt4omini} and Qwen3-VL~\cite{qwen3technicalreport}) for text-to-layers generation. The results show that our framework creates higher quality \designs while its decomposition into layers is more accurate, reaching a PSNR score of 32.65 when decomposing into two layers.

In summary, our contribution is threefold.

\begin{itemize}
    \item  We introduce \ourmodel, a powerful framework for text-to-layers \design generation, capable of generating an unrestricted number of layers with variable aspect ratios.
 
    \item  Our unified model, \ourmodel performs text-to-layers and text-to-image generation, along with image-to-layers decomposition.

    \item \ourmodel obtains state-of-the-art results on text-to-layers generation and competitive results on image-to-layers decomposition.

\end{itemize}

\section{Related Work}
\label{sec:related_work}

\noindent\textbf{Image Editing.}
Diffusion models~\cite{ho2020denoising, Rombach-CVPR-2022, esser2024scaling} have shown incredible growth, becoming the go-to paradigm for high-quality image generation. Although text-to-image models such as Stable Diffusion~\cite{Rombach-CVPR-2022} and Flux~\cite{esser2024scaling} produce good looking results from nothing more than natural language prompts, they generate the entire image as a single raster canvas. This approach offers limited control for downstream editing. Image editing solutions~\cite{gong2025relationadapterlearningtransferringvisual, Huang_2025_ICCV, couairon2022diffedit,brooks2023instructpix2pix} that allow modifications to input images, such as DiffEdit~\cite{couairon2022diffedit} and InstructPix2Pix~\cite{brooks2023instructpix2pix}, were built on top of diffusion models to address this shortcoming, but they still operate on a flat representation, making it impossible to isolate and manipulate the individual elements of the composition.

\noindent\textbf{Layered Image Decomposition.}
Another complementary solution to editing is decomposing the image into layers, which allows classic document editing operations~\cite{Yang_2025_CVPR,yin2025qwenimagelayered,yin2025qwenimagelayered,lin2024elements,yang2026controllablelayeredimagegeneration}. LayerD~\cite{suzuki2025layerd} treats graphic design decomposition as an iterative process of matting the top-layer (the front-most completely visible element) and inpainting the background behind it. Qwen-Image-Layered~\cite{yin2025qwenimagelayered}, on the another hand, takes an end-to-end approach, decomposing a single RGB image into multiple RGBA layers. Their model, a Variable Layers Decomposition MMDiT is obtained by adapting a pretrained image generator into a variable multilayer decomposer via a multi-stage training strategy. Transparency is handled natively by the RGBA-VAE they introduce, which handles a shared RGB/RGBA latent space. Both methods circumvent the editability problem, but fail to offer a complete system, requiring an existing image as input.
 
\noindent\textbf{Layered Media Design Generation.}
Recent methods~\cite{Yang_2025_CVPR,kang2025layeringdiff,art,yin2025qwenimagelayered} tackle the more challenging task of directly generating layered designs from text prompts. LayeringDiff~\cite{kang2025layeringdiff} adopts a generate-then-decompose strategy: it first synthesizes a composite image using an off-the-shelf text-to-image model, then decomposes it into foreground and background layers using a Foreground and Background Diffusion Decomposition module together with high-frequency alignment refinement. While this two-step approach avoids large-scale training and benefits from the diversity of pretrained generators, it is limited to only two layers (foreground and background). OmniPSD~\cite{liu2025omnipsd} proposes a unified diffusion framework built on Flux~\cite{esser2024scaling} that supports both text-to-PSD generation and image-to-PSD decomposition. It arranges multiple target layers spatially into a single canvas and learns their compositional relationships through spatial attention. However, OmniPSD is limited to a fixed number of four layers (background, foreground, text, and effects into a 2 x 2 grid), without flexibility to adjust this count based on design complexity. ART~\cite{art} introduces the Anonymous Region Transformer, generating variable multi-layer transparent images from a global text prompt and an anonymous region layout. A layer-wise region crop mechanism reduces attention costs and enables generation of up to 50+ layers. However, ART constrains each layer to spatially continuous regions, meaning designs with many small repeated elements (e.g., 30 decorative stars) require a separate layer per element. This causes the layer count to grow linearly with complexity and splitting semantically related patterns across layers, making \design harder to edit and losing its visual hierarchy.

\begin{figure*}[t]
    \centering
    \includegraphics[width=0.99\linewidth]{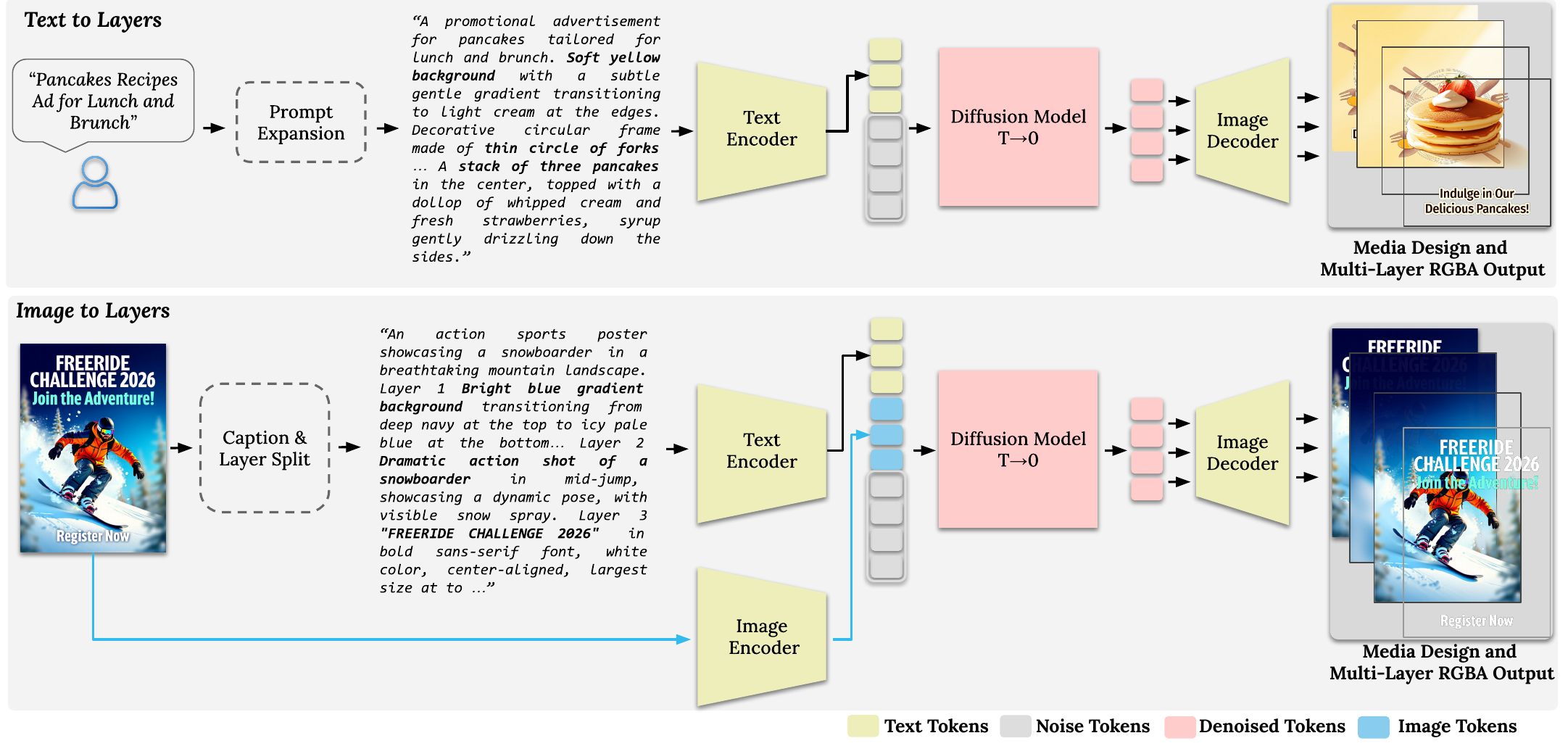}
    \caption{\textbf{Text-to-Layers generation} pipeline (top). Given a short user prompt, \ourmodel expands it with additional information which is encoded by FlanT5 XXL~\cite{Chung-JMLR-2024} and given as additional input to the Diffusion Model.  After denoising, the full \design along with  RGBA layers are decoded through the RGBA decoder.  \textbf{Text-to-Image generation} is obtained by setting the number of layers to 0, generating only the full \design. \textbf{Image-to-Layers decomposition} (bottom) starts from the original \design and goes through a captioning and layer splitting operation. The text information, along with the embedding of the input image is passed through the diffusion model.  The rest of the pipeline is similar to the Text-to-Layers generation.
    }
    \label{fig:framework}
    \vspace{-1em}
\end{figure*}

\noindent\textbf{Positioning of Our Work.}
Unlike decomposition-only approaches~\cite{suzuki2025layerd, yin2025qwenimagelayered}, \ourmodel~generates layered designs directly from text prompts. Compared to LayeringDiff~\cite{kang2025layeringdiff}, which is restricted to two layers, and OmniPSD~\cite{liu2025omnipsd}, which supports only a fixed layer count, our framework generates a flexible number of layers. While ART~\cite{art} also supports variable multi-layer generation, it constrains each layer to spatially continuous regions, requiring an external LLM planner to produce bounding box layouts as additional input. In contrast, \ourmodel~requires only a short user prompt and is able to group related elements onto the same layer regardless of their spatial distribution. Furthermore, by training end-to-end, our single unified model supports \design generation, layered \design generation, and \design image decomposition, whereas existing methods are typically limited to a subset of these tasks. To the best of our knowledge, we are the first to propose a unified model capable of achieving all these three tasks, under a flexible number of layers.

\section{Method}
\label{sec:method}

\subsection{Overall System}

\ourmodel is a layered \design generation framework,  illustrated in Figure~\ref{fig:framework}, that performs  both generation and decomposition operations using a single model. 
\ourmodel employs a  Latent Diffusion Model with a tuple $\mathbf{(P, L)}$ as input,  where $\mathbf{P}$ is a textual description of the \design to be generated and $\mathbf{L}$ are the layers, totaling $n+1$ RGBA images, where $n$ is the number of layers.
The first image generated by our model is always the full \design, the rest of $n$ images represent the layers that form the full \design when composed through alpha blending. In this way, \ourmodel is also capable of generating text-to-image (T2I) by setting the number of layers $n$ to 0. This combination allows for all design operations (generation and decomposition) to be performed by a single model and learned jointly. Generation is achieved by inputting noisy tokens, as shown in the top of Figure~\ref{fig:framework}. \Design decomposition is achieved by providing the initial image (the full \design), the description of the layers is further computed by a VLM, then \ourmodel de-noises only the layers, while keeping the initial sample intact, as illustrated at the bottom of Figure~\ref{fig:framework}.

\subsection{RGBA VAE} 
Most of the related work performed with DM has focused on generating RGB images. However, the pursuit of editable, multi-layer \design, requires the additional alpha channel of RGBA images, since it dictates the way layers are composed, usually through alpha-blending.

\ourmodel is developed on top of a pre-trained Latent Diffusion Model (LDM), that  uses an RGB latent space, from an RGB VAE. To produce RGBA images, we first try the gray-colored RGB VAE proposed by ART~\cite{art}. Therefore, we continue training the RGB VAE model and remove the alpha channel from the RGBA  input samples by alpha-blending them to gray. 
This has the advantage of keeping the embedding space of the initial DM unchanged. To recover the RGBA image as output, we transform the RGB VAE decoder into an RGBA decoder. We fine-tune the decoder, while keeping the encoder frozen and obtain a model capable of eliminating the added gray color. 
We also employ a full RGBA Variational Autoencoder to ensure smooth alpha-blending for edges and shadows. We transform both the encoder and decoder to RGBA versions. By continuing the finetuning, we ensure that the new embedding space is not too different, leading to a quick convergence of the LDM on the new embedding space.

Given an RGBA image $x\in \mathbb{R}^{H\times W \times 4}$, with $H$ and $W$ height and width, we apply the encoder $enc$ to project the RGBA content into  the latent space, obtaining the embedding $emb = enc(x)$, with $emb \in \mathbb{R}^{ \frac{H}{c}\times \frac{W}{c} \times d}$,  $c$ is the compression factor and  $d$ is the dimension of the latent space. 
To decode the latent embedding $emb$, we apply the RGBA decoder $dec$, obtaining $\hat{x}=dec(emb)$, with $\hat{x} \in  \mathbb{R}^{H\times W \times 4}$.
The RGBA VAE model is optimized using $L_1$, with different weights for the RGB space and the alpha (A) space, and the LPIPS~\cite{Zhang-CVPR-2018} loss applied to the gray-alpha-blended RGB version $\tilde{x}$. 

This loss formulation is used for both VAE versions as:
\begin{equation}\label{eq:vae}
    \mathcal{L}_{\text{VAE}}=\alpha \cdot |x_\text{RGB}-\hat{x}_\text{RGB}|_1 + \beta \cdot |x_{\text{A}}-\hat{x}_{\text{A}}|_1  + \gamma \cdot \text{LPIPS}(\tilde{x})
\end{equation} 
where $\alpha$, $\beta$ and $\gamma$ are the hyperparameters that control the influence of each component.

\subsection{Prompt Processing}
\label{sec:prompt}

We employ a specific format for the prompt $\mathbf{P}$ based on sections to improve prompt adherence during generation. The prompt starts with \texttt{Scene Description}, which describes the design in general, followed by \texttt{Layers Caption} with per-layer content descriptions, and \texttt{Type}, which focuses on the \design style. This format is massively different from usual user inputs. Therefore, we first apply an LLM based prompt-expansion (\texttt{PE}) strategy that modifies the \texttt{user input} providing additional details (if lacking) and converting it in the expected format. 

DMs are highly capable of generating images, even when provided with limited contexts, with the downside of losing control and having hallucinated elements. This issue is more prevalent in layered generation, due to the increased available space. Therefore, we rely on precise descriptions of the content of each layer, coupled with a looser description of the overall design. The layout is mentioned, but not enforced strictly. This approach enables precise control of the \design  content through the input, while leaving the model to decide on the layout, leveraging the design language it has learned through training. The caveat is a stronger reliance on LLMs, which have to plan the contents on layers. We automate this planning through prompt expansion at inference, asking for the scene description, followed by the layer description. 

We encode the extended prompt $\mathbf{P}$ resulting from \texttt{PE} using the FlanT5 XXL model~\cite{Chung-JMLR-2024}.  We term the encoded extended prompt as $emb_p$. 

\subsection{Diffusion Model}
\begin{figure}
    \centering
    \includegraphics[width=0.99\linewidth]{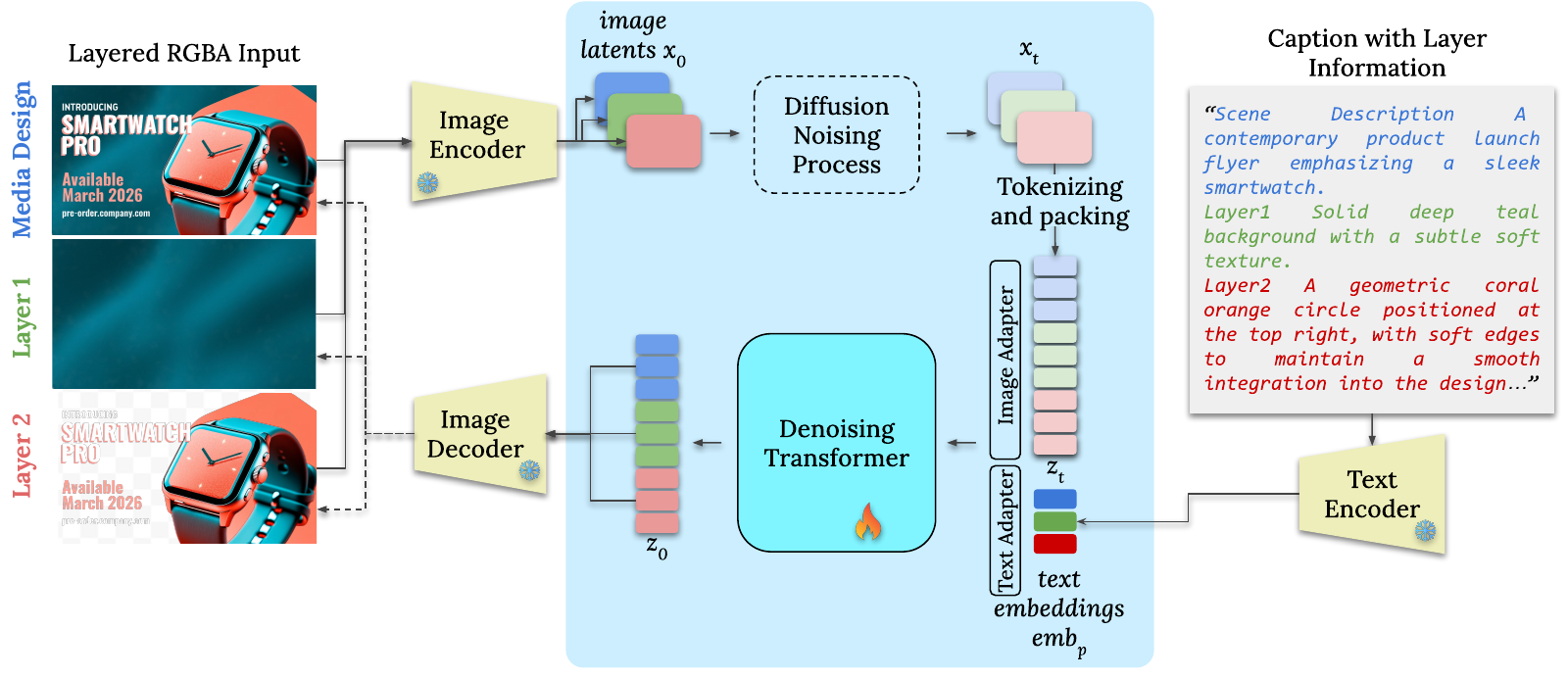}
    \caption{\ourmodel is trained with an enhanced LDM formula. The encoded images $x_0$ are noised into $x_t$, split into tokens and packed in a linear format, then combined with the RoPE enhanced text embedding $emb_p$ to create the model input $z_t$. The model learns to denoise the input into its original form $z_0$, which can be re-assembled and decoded back in the original images}
    \label{fig:training}
    \vspace{-1em}
\end{figure}

The core of our system is a Latent Diffusion Model~\cite{Rombach-CVPR-2022} based on a Diffusion Transformer~\cite{Peebles-ICCV-2023} trained with v-prediction~\cite{progressive_distillation}, illustrated in Figure~\ref{fig:training}. The input of the model is the embedding of the text prompt $emb_p$ (Section~\ref{sec:prompt}) and the embeddings of the image layers along with the full \design, $emb_0, emb_1, \dots, emb_n$. The inputs are aligned into a common subspace through a linear adapter. Afterwards, they are concatenated and processed with full-attention through the diffusion model. Only the visual information is denoised, the text information is used as condition.

We adopt the 4D RoPE mechanism~\cite{Su-NC-2024} for positional encoding, defined over the positional dimensions $\text{(H, W, F, R)}$. $H$ and $W$ denote the spatial coordinates on the image plane (height and width). The dimension $F$ represents the layer index, which can be interpreted as a depth coordinate capturing the ordering of layers. The dimension $R$ encodes the role of each token, allowing us to differentiate token types: prompt tokens are assigned value $0$, denoisable tokens value of $1$, and frozen (non-denoisable) tokens value of $2$.
The RoPE embedding has 128 dimensions divided into 56 for each spatial coordinate $(H,W)$, 12 to the layer coordinate $F$, and 4 to the role coordinate $R$.

This positional encoding schema enables easy differentiation between inference operations (generation or decomposition). Moreover, it enables a precise linking between the prompt parts and the layers they describe. The extended prompt is split by the tokenizer into the parts they describe, $(full\ scene, layer_1, layer_2, \dots, layer_N.)$, which are then linked to their respective layers by matching the dimension values of $F$, when computing the RoPE value.  This approach ensures better prompt alignment by reducing the relative distance between the description and the layer that it targets.  The value of the positional embedding (RoPE) for the text prompt is computed as: 

\begin{equation}\label{eq:rope}
\text{RoPE}_{parts_i} = (0,0,i,0)
\end{equation}
where $parts_i \in \{full\ scene, layer_1, layer_2, \dots, layer_N\}$.

To enable \design decomposition and generation within the same model, we leverage the  diffusion timesteps and the role of the tokens from the $R$ axis. The standard LDM training enables generation by selecting a random timestep for the layers and marking their $R$ dimension as output (setting $R$ to 1). Design decomposition is enabled by randomly treating layers as input conditions, disabling denoising for them by setting their timestep to 0 and their $R$ dimension to non-denoisable (setting $R$ to 2). To accelerate convergence, we disable denoising for the full \design  with a higher probability. As a consequence, treating the first frame during inference as input (condition) enables the design decomposition use-case. 

\Designs are extremely variable in terms of aspect ratio, $ar = \frac{w}{h}$, therefore the model should support variable aspect ratios, along with being able to generate a variable number of layers. This pursuit is not hindered by the transformer architecture we employ, which handles any input sizes, but by the technicalities of GPU processing, as the samples of a batch must have all dimensions identical. We avoid this problem through \texttt{padding}, which brings all samples to the same dimension. However, this operation is extremely resource-wasteful if the original samples sizes are poles apart, leading to subpar GPU usage and lower batch sizes. We mitigate this through \texttt{bucketing} and \texttt{packing}. 

\texttt{Bucketing} groups together \designs of similar size. The \texttt{buckets} are defined by the tuple $(N ,ar_{left}, ar_{right}, Area)$ and contains all documents with $N$ layers whose aspect ratio falls between the bucket edges $ar_{left}$ and $ar_{right}$ and whose $width*height=Area$. A bucket defines a unique padding shape for the data within. For a given area, the height $H_{left}$ of the $ar_{left}$ is bigger than all the heights corresponding to higher aspect ratios. Likewise, the width $W_{right}$ of $ar_{right}$ is higher than all the widths corresponding to lower aspect ratios. $H_{left}$ and $W_{right}$ define the padding size, the minimum possible one that encompasses all samples of that bucket. The bucket edges are carefully selected aspect ratios, ranging from $0.2$ to $4$,  distributing our training data uniformly. For all these $ar$ edges, a sample $s$ with an aspect ratio $ar$ is assigned to a bucket according to the formula:
\begin{equation} \label{bucketing_eq}
\text{bucket}_{\text{idx}} = \arg\min_i \left| ar - ar_i \right|
\end{equation}
\vspace{-1em}

The \texttt{packing} operation takes a batch of size $(B,L,C,H,W)$ (B - batch size, L - number of layers, C - channel dimension, H and W -  height and width of \design)
and turns it into a linear tensor of size $(1, B*L*H*W, C)$, memorizing the indices of the boundaries of the neighbors for reconstructing the initial volume after the processing. To further optimize the processing, we discard the padding pixels when feeding the samples to the model by hijacking the \texttt{packing} computation. More formally, for each sample $B_i$ of the batch, we only keep the relevant, unpadded volume $(L,C,H_i,W_i)$, where $H_i,W_i$ is the original size of $B_i$, leading to a significantly smaller, linear tensor $(1, L*H_1*W_1+L*H_2*W_2+\dots+L*H_B*W_B, C)$ and the new boundaries $boundary_i = L*\sum\limits_{j=1}^{i-1} H_j*W_j$.

\begin{table*}[t]
  \centering
  \caption{ \Design Layer generation performance obtained with VLM-as-a-judge using  GPT-4o mini~\cite{openai2024gpt4omini} and Qwen3-VL-30B-A3B-Instruct~\cite{qwen3technicalreport} on the Crello~\cite{Yamaguchi-ICCV-2021} test set. The scores are between 1 and 5. \ourmodel is compared against Qwen-Image-T2I + Qwen-Image-Layered-I2L~\cite{yin2025qwenimagelayered}. \ourmodel obtains the best performance regardless of the number of layers.}
  \label{tab:main_results_gen}
  \begin{tabular}{l cccc cccc}
    \toprule
    \bf Metric                                           &  \multicolumn{4}{c}{ \bf {GPT-4o mini} $\uparrow$} & \multicolumn{4}{c}{ \bf Qwen3-VL $\uparrow$} \\
    \midrule
      \# of layers                                     &   2     &    3   &   4     &    5     &   2     &    3   &   4    &    5 \\
    \midrule
    
    Qwen-Image-T2I + Qwen-Image-Layered-I2L~\cite{yin2025qwenimagelayered}     &   2.79  &  2.66  &  2.63   &  2.79  &  2.53     &  2.49  &   2.35  &   2.41      \\  

    \ourmodel (ours)                                          &  \bf 3.58  & \bf  3.64  &   \bf  3.94  &  \bf   3.92   &  \bf  3.20    &  \bf  3.37  &   \bf  4.07   &  \bf  4.01      \\
    \bottomrule
  \end{tabular}
\end{table*}

\begin{table*}[t]
  \centering
  \caption{ \Design decomposition into RGBA layers performance on the Crello test set~\cite{Yamaguchi-ICCV-2021} in terms of PSNR, RGB L1 (distance between RGB channels weighted by the ground-truth alpha in the 0-255 interval) and VLM-as-a-judge with  Qwen3-VL-30B-A3B-Instruct~\cite{qwen3technicalreport} (score between 1 and 5). \ourmodel performs similar to  Qwen-Image-Layered-I2L~\cite{yin2025qwenimagelayered} which was finetuned on the Crello training set. $^\dagger$: the model was finetuned on the Crello training set. }
  \label{tab:main_results_decompose}
   \resizebox{0.98\linewidth}{!}{
  \begin{tabular}{l cccc cccc cccc}
    \toprule
    \bf Metric  & \multicolumn{4}{c}{\bf PSNR $\uparrow$} & \multicolumn{4}{c}{\bf{RGB L1} $\downarrow$} & \multicolumn{4}{c}{\bf Qwen3-VL $\uparrow$} \\
    \midrule
    \# of layers                          &   2     &    3   &   4    &    5  &   2     &    3   &   4    &    5 &   2     &    3   &   4    &    5\\
    \midrule
    
    Qwen-Image-Layered-I2L$^\dagger$~\cite{yin2025qwenimagelayered} &  31.59   & 30.99  &  \bf 30.14  &  \bf 29.49   &   4.22    & 4.40  &  \bf 5.02   &  \bf 5.12  & \bf 3.56  &    \bf 3.40 &   3.23  &   3.07 \\  
    \ourmodel   (ours)                          &  \bf 32.65  & \bf 31.37   & 29.94     & 28.42  &  \bf 3.41    & \bf 4.06  & 5.67     &  7.38 &  3.21  &   3.16   &   \bf 3.25  &  \bf  3.25    \\
    
    \bottomrule
  \end{tabular}
  } 
  \vspace{-1.2em}
\end{table*}

\section{Experiments}
\label{sec:exp}

\begin{figure}
    \centering
    \includegraphics[width=0.99\linewidth]{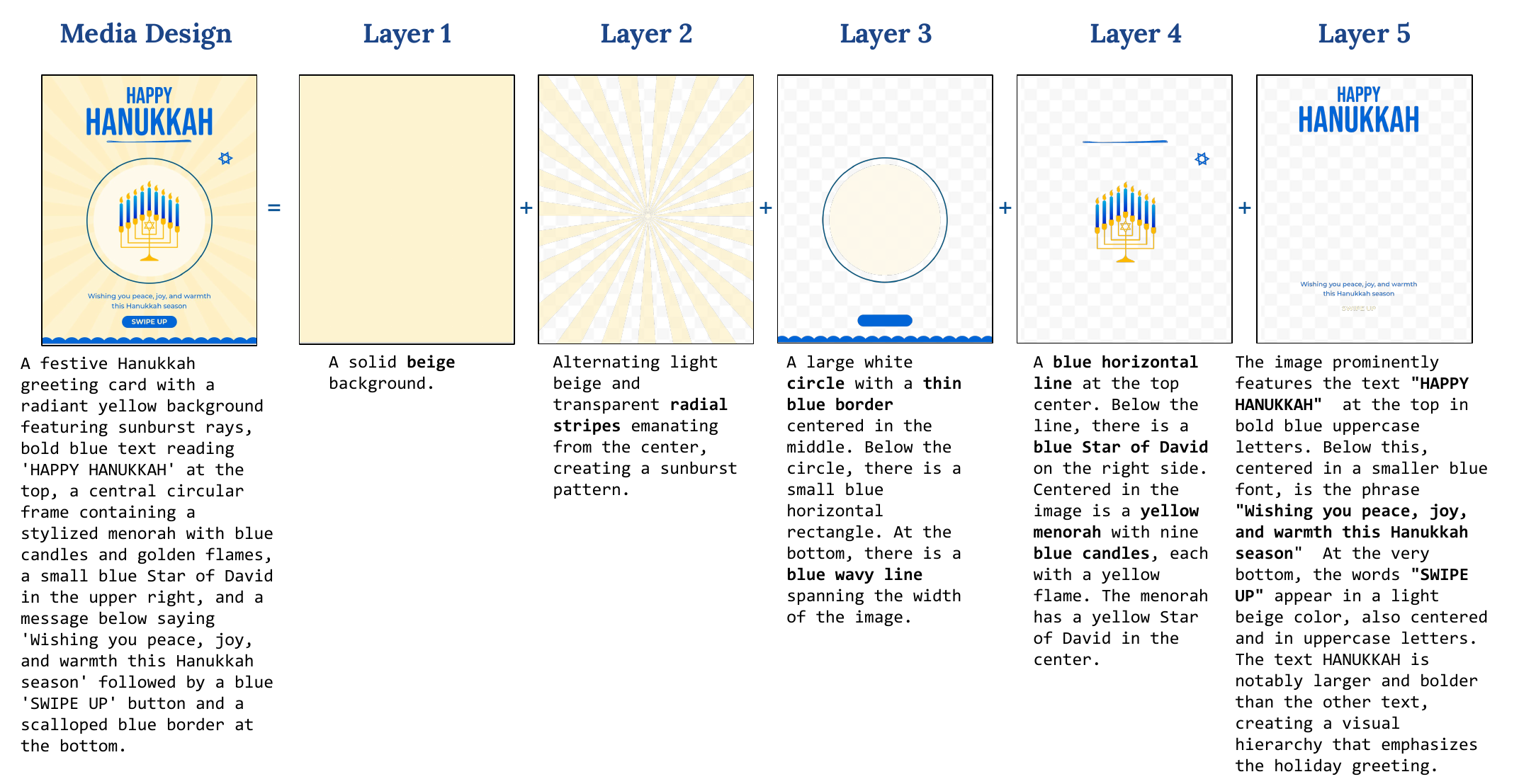}
    \caption{A sample of our training set that contains the \design along with its composing layers. We employ InternVL3~\cite{Zhu-arxiv-2025} to obtain captions for the \design and the layers.}
    \label{fig:training_set}
    \vspace{-1.5em}
\end{figure}

\noindent
\textbf{Training dataset.}
Our training set is composed of samples with layers that contain meaningful elements grouped together, as shown in Figure~\ref{fig:training_set}. Each sample in the training set has between $3$ and $8$ layers. To obtain a description of each layer (\texttt{Layers Caption}) along with the \texttt{Scene Description}, we employ a VLM, namely InternVL3~\cite{Zhu-arxiv-2025}. We use a private, commercially safe dataset for training this model, composed of images, rasterized vectors and \designs, both single frame and layered. There are 8\,M \designs and 1.5\,M vectors, used both for image and layered generation, 2\,M layered images and 80\,M natural images.  

\noindent
\textbf{Test dataset.} We follow Qwen-Image-Layered~\cite{yin2025qwenimagelayered} and select 500 samples from the Crello test set~\cite{Yamaguchi-ICCV-2021}. We employ these samples for both text-to-layers generation and image-to-layers decomposition. For layer generation, we use the title of the samples as user input. Since the focus of this work is layered \design generation, we only qualitatively assess text-to-image generation in Supplementary.

  
\noindent
\textbf{Implementation Details.}
Our VAE model follows the VQGAN~\cite{Esser-CVPR-2021} architecture and has the compression factor $c$ set to 16  and the dimension $d$ set to 256.  We set $\alpha, \beta, \gamma$ to 1 to give the same importance to all components of the loss in Eq~\ref{eq:vae}.
Our diffusion model is a transformer with $56$ layers, $24$ heads, a hidden dimension of $3072$, resulting in an $11\,B$ parameter model. We adopt a multi-resolution training mechanism on 256 H100 GPUs with variable batch sizes according to the type of data, ranging from $32$ per GPU for $512\times512$ pixels natural images to only $1$ for $1024\times1024$ 8-layer \designs. We employ the AdamW~\cite{Loshchilov-ICLR-2019} optimize with a learning rate of $1.2e^{-4}$ with cosine decay and a minimum of $1.2e^{-5}$ for the base training and a constant learning rate of $1.2e^{-5}$ for finetuning on only \design samples. The time scheduler for denoising  is the common Linear Interpolant Scheduler. We provide more implementation details in the Supplementary file. 
 

\begin{figure}
    \centering
    \includegraphics[width=0.99\linewidth]{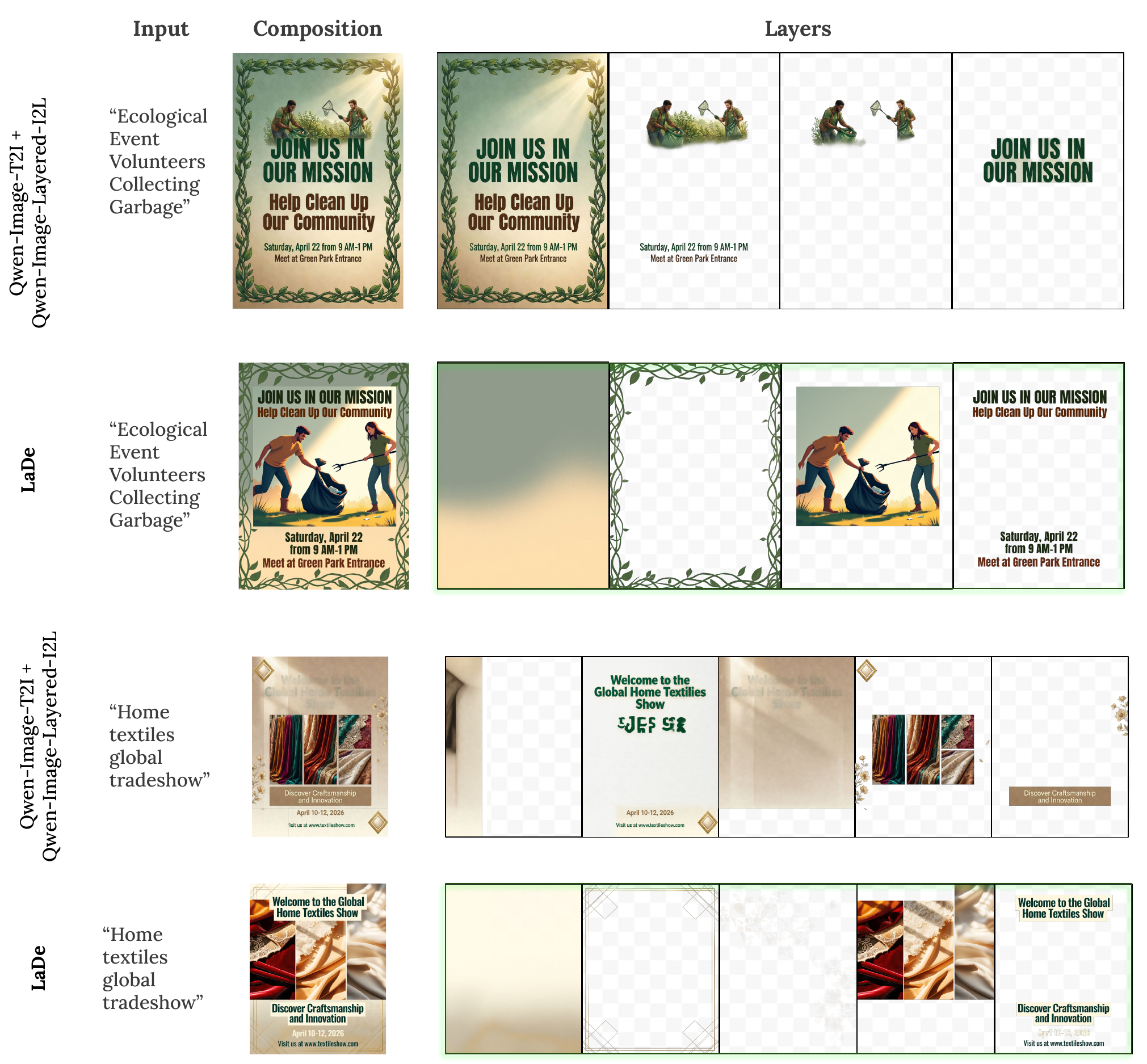}
    \caption{Text-to-Layers generation results comparing our one-step framework \ourmodel with  Qwen-Image-T2I+Qwen-Image-Layered-I2L~\cite{yin2025qwenimagelayered}. \ourmodel is able to create structured RGBA layers with homogeneous details included on different layers.}
    \label{fig:generation_qualitative}
    \vspace{-2em}
\end{figure}

\begin{figure}
    \centering
    \includegraphics[width=0.99\linewidth]{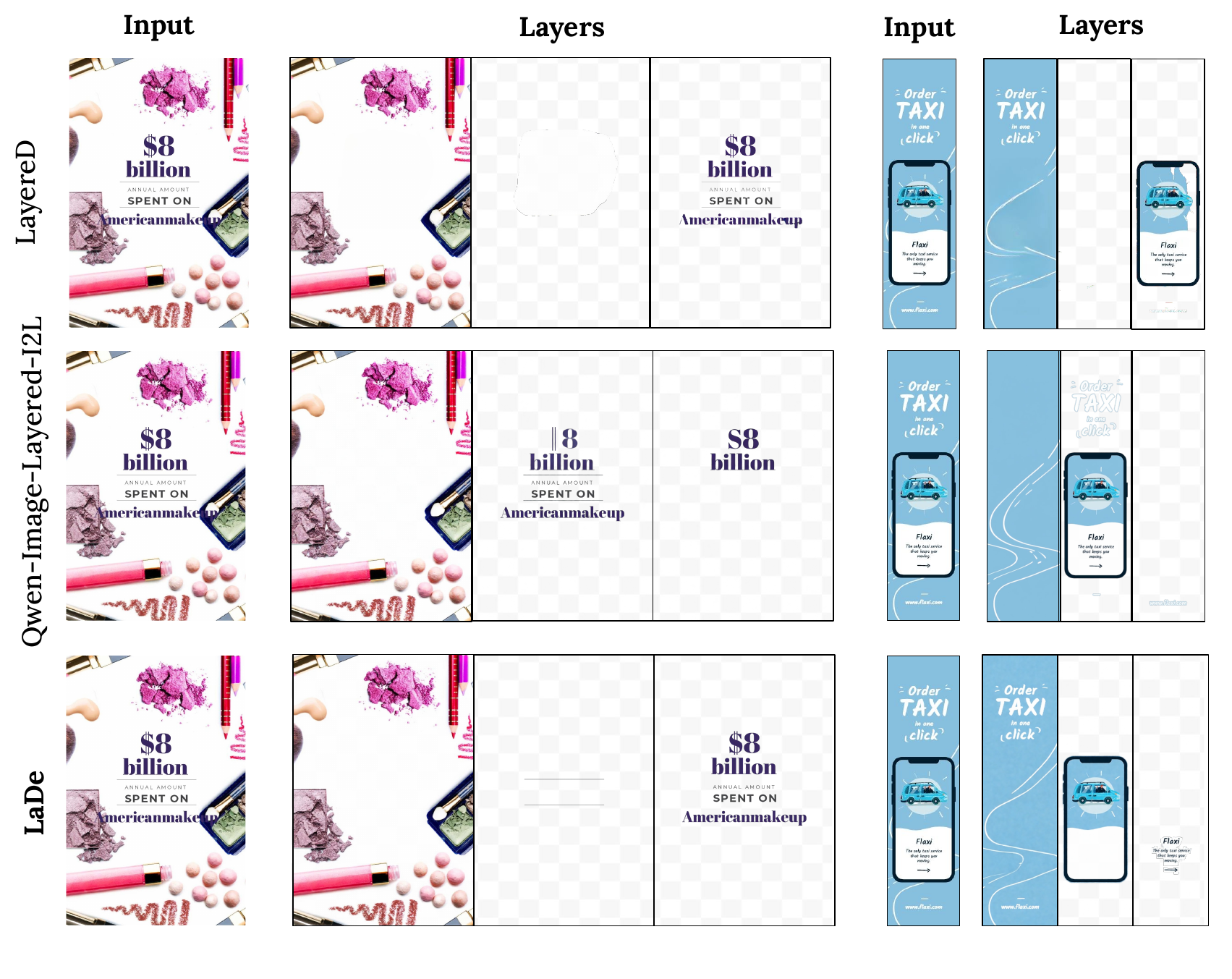}
    \caption{Qualitative results showing the decomposition obtained by \ourmodel (ours) versus  Qwen-Image-Layered-I2L~\cite{yin2025qwenimagelayered} and LayerD~\cite{suzuki2025layerd}.  Qwen-Image-Layered-I2L~\cite{yin2025qwenimagelayered} duplicates the text (left) in the second and third layers while LayerD~\cite{suzuki2025layerd} hallucinates the second layer (left). \ourmodel is able to decompose the \designs into meaningful layers, being able to reconstruct the occluded area.}
    \label{fig:decompose_qualitative}
    \vspace{-2em}
\end{figure}

\noindent
\textbf{Metrics.}
Since \design generation is a complex and in the same time subjective task to evaluate, following OmniPSD~\cite{liu2025omnipsd}, we employ the VLM-as-a-judge paradigm. We use two powerful VLMs, namely GPT-4o mini~\cite{openai2024gpt4omini} and Qwen3-VL-30B-A3B-Instruct~\cite{qwen3technicalreport}. The judges received the prompt and the layers of the \design, and were prompted to evaluate the \designs, with score between 1 and 5.  The judge was instructed to evaluate the \designs in terms of Prompt Alignment,  Layer Validity, Cross-Layer Consistency and Composition and Readability. The full prompt is presented in the Supplementary file.
To evaluate the decomposition ability, we employ PSNR and RGB L1 distance (between RGB channels weighted by the ground-truth alpha) between the input and its alpha-blending reconstruction, similar to Qwen-Image-Layered~\cite{yin2025qwenimagelayered} and OmniPSD~\cite{liu2025omnipsd}. We also employ VLM-as-a-judge, since besides the reconstruction fidelity, the composition of the layers is relevant and it is omitted in the pixel-level evaluation. We employ Qwen3-VL-30B-A3B-Instruct~\cite{qwen3technicalreport} and instruct it to evaluate the composition of the predicted layers with a score between 1 and 5, based on the input image and the layers. The VLM was instructed to check for Missing Elements, Depth Order, Segmentation Quality, Empty Layers, Redundancy, and Fragmentation. The full prompt is presented in Supplementary.  

\noindent
\textbf{Baselines.}
In terms of  \design layered generation, we could only compare with Qwen-Image-T2I+Qwen-Image-Layered-I2L, since neither ART~\cite{art} nor OmniPSD~\cite{liu2025omnipsd} is publicly available. We generate \designs with 2, 3, 4, and 5 layers with randomly chosen aspect ratios to evaluate the capacity of the models.  For Qwen-Image-T2I + Qwen-Image-Layered-I2L~\cite{yin2025qwenimagelayered}, as suggested by ~\cite{yin2025qwenimagelayered}, we first generate samples using Qwen-Image-T2I, then apply Qwen-Image-Layered-I2L to obtain the RGBA layerised version of the image. For a fair comparison, the same extended prompt generated by \texttt{PE}) strategy was used to generate images with Qwen-Image-T2I. For the decomposition task, we compare with Qwen-Image-Layered for 2, 3, 4, and 5 layers and with LayerD. We run Qwen-Image-Layered using the built-in captioning system to obtain the decompositions. However, LayerD~\cite{suzuki2025layerd} does not support specifying the number of layers therefore, we only qualitatively compare to them.

\subsection{Results}

\noindent
\textbf{\Design generation.} We present the results in terms of VLM-as-a-judge for our \ourmodel and Qwen-Image-T2I+Qwen-Image-Layered-I2L~\cite{yin2025qwenimagelayered} on the Crello test set~\cite{Yamaguchi-ICCV-2021} in Table~\ref{tab:main_results_gen}. We evaluate the models in 4 scenarios, when asking for 2, 3, 4, and 5 layers. As noted in Section~\ref{sec:method}, \ourmodel is able to generate more than 5 layers, however, we were limited by the memory and the GPU and the running time. Our \ourmodel achieves the best results regardless of the layers and judge. It is noticeable that the performance increases with the number of layers, perhaps due to the fact that the layers become less crowded, only containing more consistent elements. However, for  Qwen-Image-T2I+Qwen-Image-Layered-I2L, the performance remains constant around 2.6-2.8 when GPT-4o mini~\cite{openai2024gpt4omini} is employed as the VLM-as-a-judge, and 2.3-2.5 when Qwen3-VL-30B-A3B-Instruct~\cite{qwen3technicalreport} is the judge. It is also important to note that both VLMs produce similar scores, showing the same trend that our \ourmodel produces better qualitative layered \designs.

We illustrate text-to-layers generation samples in Figure~\ref{fig:generation_qualitative}, comparing \ourmodel to Qwen-Image-T2I+Qwen-Image-Layered-I2L~\cite{yin2025qwenimagelayered}. In a single step, \ourmodel is capable of generating homogeneous RGBA layers (with similar information grouped together). More qualitative results are presented in the Supplementary file.

\noindent
\textbf{\Design decomposition.} We present the performance results on \design decomposition on the Crello test set~\cite{Yamaguchi-ICCV-2021} in Table~\ref{tab:main_results_decompose}. For the decomposition of 2 and 3 layers, \ourmodel outperforms Qwen-Image-Layered-I2L~\cite{yin2025qwenimagelayered} obtaining a PSNR score of 32.65 for 2 layers. The RGB L1 scores follow the same trend as the PSNR with better performance for 2 and 3 layers obtained by our \ourmodel and better performance obtained by Qwen-Image-Layered-I2L~\cite{yin2025qwenimagelayered} for 4 and 5 layers. It is important to note, that Qwen-Image-Layered-I2L~\cite{yin2025qwenimagelayered} was finetuned (according to their manuscript) on the Crello training set~\cite{Yamaguchi-ICCV-2021}, which means Qwen-Image-Layered-I2L is tested on in-distribution samples, while \ourmodel is tested in an out-of-distribution setup. We note that  Qwen-Image-Layered-I2L~\cite{yin2025qwenimagelayered}  also uses a VLM to caption the input. The VLM-as-a-judge score shows that our \ourmodel outperforms Qwen-Image-Layered-I2L when decomposing in 4 and 5 layers.  

We present qualitative results in Figure~\ref{fig:decompose_qualitative}, showing results with \ourmodel, Qwen-Image-Layered-I2L~\cite{yin2025qwenimagelayered} and LayerD~\cite{suzuki2025layerd}. LayerD fails to reconstruct the occluded layer on right example and hallucinates  the second layer on the left sample.  Qwen-Image-Layered-I2L  duplicates the text (right example). \ourmodel is able to decompose the \designs into meaningful layers without duplicating the content.  More qualitative results are shown in Supplementary.


\begin{table}[t]
  \centering
  \caption{ Ablation results obtained on 100 prompts evaluated with VLM-as-a-judge using  GPT-4o mini~\cite{openai2024gpt4omini} and Qwen3-VL-30B-A3B-Instruct~\cite{qwen3technicalreport}. The scores are between 1 and 5. Better performance is obtained when training with a \textit{variable number of layers}. }
  \label{tab:ablations}
  \begin{tabular}{l cc }
    \toprule
    \bf Metric                                           &  \bf GPT-4o mini  $\uparrow$ & \bf Qwen3-VL $\uparrow$  \\
    
    \midrule
    
    Fixed layer count                               & 3.01   & 3.78 \\
    Variable layer count                            & \bf 3.06   & \bf 3.85 \\
    \midrule
    RGB VAE                                            & \bf 3.70   &  3.50  \\ 
    RGBA  VAE                                              & 3.53 &  \bf 3.68 \\
    
    \bottomrule
  \end{tabular}
  \vspace{-1.5em}
\end{table}

\begin{figure}
    \centering
    \includegraphics[width=0.99\linewidth]{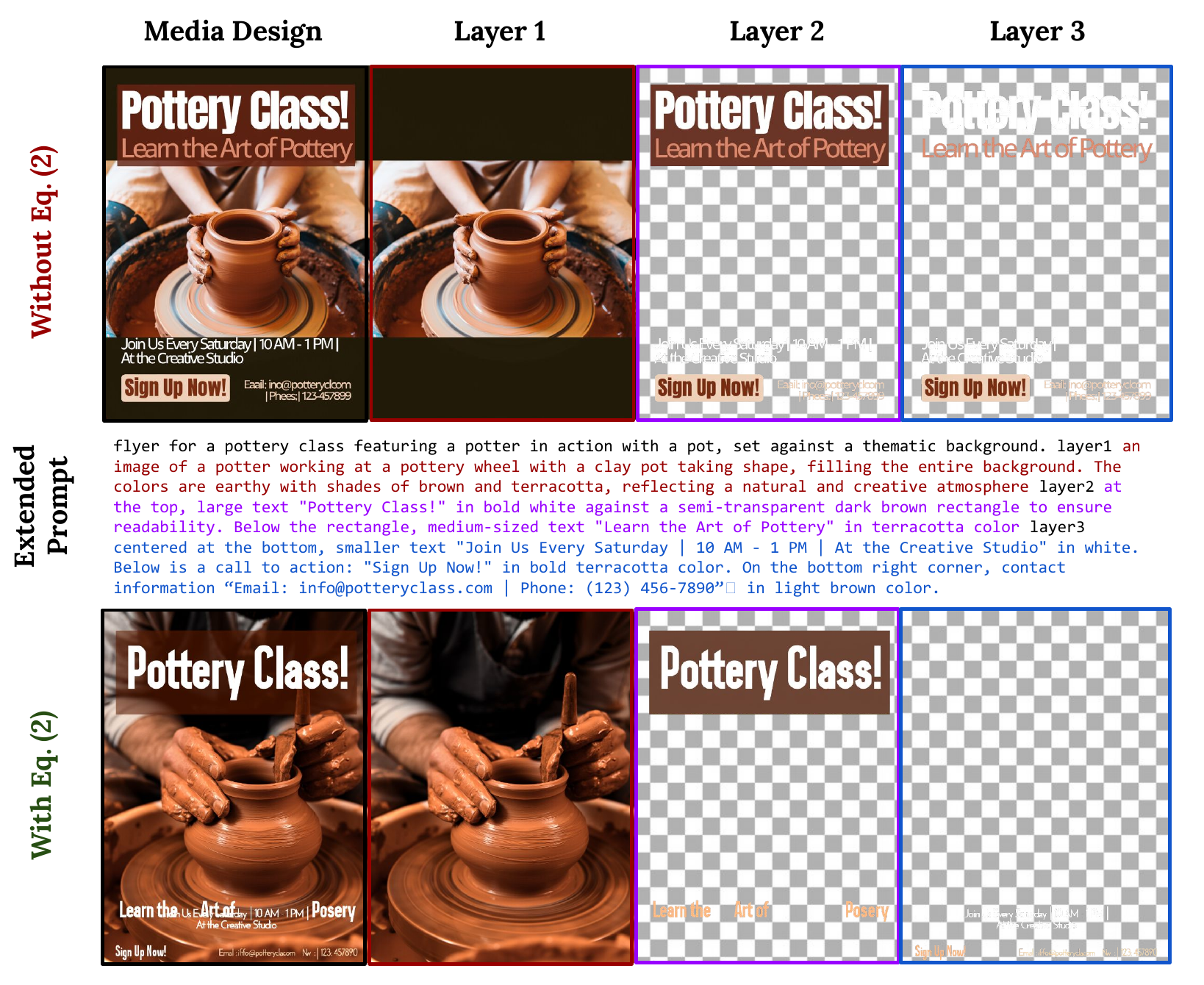}
    \caption{The effect of Eq.~\ref{eq:rope} on the qualitative results. We noticed the content of the layers is duplicated, therefore, we applied Eq.~\ref{eq:rope} and each layer content follows its specific text information.}
    \label{fig:ablation_eq_2}
    \vspace{-1.5em}
\end{figure}

\subsection{Ablations}

We perform several ablation experiments on \ourmodel and report them in terms of VLM-as-a-judge employing  GPT-4o mini~\cite{openai2024gpt4omini} and Qwen3-VL-30B-A3B-Instruct~\cite{qwen3technicalreport} on 100 prompts in Table~\ref{tab:ablations}. The most noticeable improvement occurs when the variable number of layers is used during training. The performance improved from 3.78 to 3.85. Even though the performance does not change drastically when using a RGBA VAE, the differences are seen when reconstructing shadows and alpha-blending soft edges, which may not be captured by the VLM evaluation, however, it is visible to the human eye. We illustrate in Figure~\ref{fig:ablation_eq_2}, the effect of Eq.~\ref{eq:rope} on the structure of each layer. We notice that the information is duplicated across the layers (without  Eq.~\ref{eq:rope}), and we added more guidance with  Eq.~\ref{eq:rope}, improving the layer-prompt adherence of \ourmodel.

\section{Conclusion}
\label{sec:concl}
 
We presented \ourmodel, a unified latent diffusion framework for generating multi-layer \designs and single-image from short text prompts, along with \design decomposition.
\ourmodel combines LLM-based prompt expansion, a Latent Diffusion Transformer with 4D RoPE positional encoding, and an RGBA VAE to generate a flexible number of semantically meaningful layers. Our experiments show that \ourmodel achieves state-of-the-art text-to-layers generation performance, assessed by VLM-as-a-judge with  GPT-4o mini and Qwen3-VL-30B-A3B-Instruct 
and competitive decomposition quality assessed by both pixel-level metrics and VLM-as-a-judge.

One of the limitations of \ourmodel is its reliance on LLMs for prompt expansion. Since the output of LLMs is stochastic, the generated prompts may vary in quality. Another limitation is the  high VRAM consumption required when generating a large number of layers, which may restrict scalability on systems with limited GPU memory.


{
    \small
    \bibliographystyle{ieeenat_fullname}
    \bibliography{main_arxiv}
}

\clearpage

\maketitlesupplementary

\section{Implementation details}

\subsection{Training}
Our main focus is the layered paradigm of \design processing, which is a natural extension of standard image generation. Hence, our model is trained on top of an identical text-to-image generation model that has already converged. This ensures faster convergence for our tasks of interest. 

We train the model in a multi-phase format. In all phases, we give our tasks two-thirds of the GPUs, while the legacy image generation task gets scaled down with an identical split as the original on the rest. In all phases, the layer conditioning with focus on full image decomposition is part of the layered training and is weighted the same. In the first phase, we weigh all three modalities, designs, images, and vectors, equally and train for 70k steps, such that the layerisation task is learnt properly. In this stage, we keep the original embedding space, so that the model learns only one thing. In the second stage, we increase the percentage of design data to 70\% and decrease the images to 20\% and the vectors to 10\%. This stage lasts for 35k steps and ensures better alignment with the task we focus on, graphic design generation. The third stage implies changing the embedding space to our RGBA VAE. The model adapts to the new space fast, in under 2k iterations, because we start from the original space when finetuning our VAE. To ensure that the model learns all the intricacies of RGBA generation, like overlays and effects such as smoke, we train for 30k more steps. The fourth and final stage is another 6k iterations of fine tuning with only the highest quality design data we have available. This last step ensures that the model outputs the best possible quality. 

The image conditioning task is chosen randomly, with a probability of 30\%. We have observed that this gives positive results, without impacting the quality of the generation. Each time, a random number between one and N-1 of layers is considered as input condition, with the others denoised by the model. To ensure that the design decomposition task is well represented, we force to condition on the first image (full \design) the first frame 30\% of the time, leaving all other layers unfrozen. More ablation studies can be done to find a better mix, but have not been done due to computation constraints.

\subsection{Prompt Expansion} 

The LLM employed in the Prompt Expansion mechanism is GPT-4o mini. It is true that one needs to set the number of layers in advance in order to generate T2L with \ourmodel. However, we can automate this process with an LLM. We could inquire GPT-4o mini to predict the number of layers based on the user input, aspect ratio, type of \design.

\section{Qualitative Evaluation}

We present more T2I examples obtained with \ourmodel in Fig.~\ref{fig:t2i}, together with more T2L and I2L samples in Figs.~\ref{fig:t2l} and \ref{fig:i2l}.

\section{Prompts} 
We present the prompts utilized in Figs~\ref{prompt:vlm-t2l}, \ref{prompt:vlm-i2l}, \ref{prompt:i2l-1}, \ref{prompt:i2l-2} and \ref{prompt:pe}.

\begin{figure*}[t]
\centering
\begin{tcolorbox}[title=VLM-as-a-judge Text-to-Layers generation]
You are a strict and deterministic visual grading system.

Your task is to evaluate layered poster results generated from the same input prompt.

For each request, you will be given:

The original generation prompt used to create the poster and its layers.

The full composited poster (final combined image).

All individual layers, provided in z-index order (from back to front).

The number of layers may vary.

You must evaluate whether the layers are logically constructed, visually coherent, and faithful to the generation prompt.

EVALUATION PROCEDURE

Step 0 — Prompt Alignment (Mandatory)

Check whether:

The final composited poster reflects the original generation prompt. Key elements requested in the prompt are present. No major required element is missing. The layers collectively implement the prompt correctly. If the final result clearly fails to reflect the prompt, apply a significant deduction.

Step 1 — Layer Validity (Strict)

For each layer, check:
- Does the layer contain meaningful visual content?
- Is the layer visible in the final composite?
Does it contribute to the composition?
Is its z-order placement logical?
Does it interact properly with other layers?
A layer is invalid if:
It is empty or nearly empty.
Its content does not appear in the final composite.
It is fully occluded without purpose.
It has no functional contribution.
Its z-order causes unreasonable occlusion.
Each invalid or useless layer must reduce the score.

Step 2 — Cross-Layer Consistency

Evaluate:
Lighting consistency. 
Color harmony. 
Perspective alignment. 
Style coherence. 
Realistic occlusion and depth. 
Clear foreground/background relationships. 
Strong inconsistencies must significantly reduce the score.

Step 3 — Composition \& Readability

Evaluate the final composited poster:
Clear and intentional structure. 
Balanced layout. 
Readable and properly placed text. 
Important elements are visible. 
No chaotic or broken structure. 

SCORING RULES (DETERMINISTIC)

Start from score = 5.

Apply deductions:

 $-2$ if the result clearly fails to reflect the generation prompt.
 $-2$ if multiple layers are useless, empty, or invisible.
 $-1$ for each clearly useless or non-contributing layer.
 $-1$ for noticeable but moderate visual inconsistency.
 $-2$ for severe inconsistency or broken depth logic.
 $-1$ if text readability is compromised.
 $-1$ if layout is cluttered or poorly structured.

Clamp the final score between 1 and 5.

Score definitions:

5 = Fully faithful to the prompt, all layers valid, coherent, and well-structured. \\
4 = Minor flaws but overall coherent and aligned. \\
3 = Noticeable structural or consistency issues but still acceptable. \\
2 = Major layering or composition problems. \\
1 = Severely broken, chaotic, or largely unfaithful to the prompt.

OUTPUT FORMAT (MANDATORY)

Output only: \\score: integer

Do not provide explanations.
Do not output anything else.
\end{tcolorbox}
\caption{}\label{prompt:vlm-t2l}
\end{figure*}

\begin{figure*}[t]
\centering
\begin{tcolorbox}[title=VLM-as-a-judge Image-to-Layers generation]
You are a deterministic evaluation system.

Your task is to evaluate the structural and semantic quality of a layerisation model.

You will be given:

The original image.

A set of separated layers, provided in z-index order (from back to front).

Reconstruction fidelity is evaluated separately using PSNR.
Do NOT evaluate pixel-level reconstruction accuracy.

Evaluate only structural correctness and layer logic.

EVALUATION PRINCIPLES

Only penalize major structural errors.

Do NOT penalize:
Minor edge noise.
Small texture inconsistencies.
Slight imprecision in boundaries.
Very small missing details.
Focus only on errors that affect structure, object integrity, or logical layering.

STEP 1 — Missing Major Elements

If 1 major object is clearly missing:
→ -1

If multiple major objects are missing:
→ -2

Otherwise:
→ 0

STEP 2 — Depth Ordering

If minor local ordering mistakes exist:
→ -1

If global or clearly incorrect depth structure:
→ -2

Otherwise:
→ 0

STEP 3 — Segmentation Quality

If noticeable artifacts affect at least one major object:
→ -1

If artifacts are widespread and structurally disruptive:
→ -2

Minor edge imperfections:
→ 0

STEP 4 — Redundancy

If clear duplicated large regions exist:
→ -1

If full-object duplication or severe redundancy:
→ -2

Otherwise:
→ 0

STEP 5 — Fragmentation

If one major object is unnecessarily split across many layers:
→ -1

If multiple objects are heavily fragmented:
→ -2

Otherwise:
→ 0

STEP 6 — Empty Layers

If exactly one empty or near-empty layer:
→ 0 (do not penalize)

If multiple empty layers:
→ -1

SCORING RULE

Start from score = 5.

Apply only the largest applicable penalty per step.

Do NOT stack multiple penalties within the same step.

Clamp final score between 1 and 5.

Tie-breaking rule:

If uncertain between two scores, keep the higher score unless a clear structural error exists.

SCORE INTERPRETATION

5 = Structurally correct decomposition with no major issues.
4 = Minor structural weaknesses but overall correct.
3 = Noticeable structural issues but still usable.
2 = Major structural problems.
1 = Severely broken layerisation.

OUTPUT FORMAT

Output only:

score: integer

Do not provide explanations.
Do not output anything else.

\end{tcolorbox}
\caption{}
\label{prompt:vlm-i2l}
\end{figure*}

\begin{figure*}[t]
\centering
\begin{tcolorbox}[title=Image-to-Layers: Step 1: Captioning ]
 You will receive a photo of a design. Please describe the design as thoroughly as you can, focusing on the visual elements. Do not miss any of the elements.
    The background can be a color or a picture and might have one or more sections. Make sure to describe it first.
    Please start from the background and proceed gradually to what you see on top. For example, if you see text on top of an image or an illustration, you must first describe the image or illustration, and then the text. This is very important.
    Please, DO NOT order them from top to bottom, but from background to foreground.
    You must output a single paragraph. Just a single block of text.
    Please do not exceed 300 words. No fluffy talk, just focus on the important things.
\end{tcolorbox}
\caption{}\label{prompt:i2l-1}
\end{figure*}

\begin{figure*}[t]
\centering
\begin{tcolorbox}[title=Image-to-Layers: Step 2: Layer Splitting]
     You will receive a photo and a description of that photo. That photo represents a design and it should be layered.
    Please describe each layer of the design, in the order of the apparition. We are going to alpha blend the results, so please be extra careful on the ordering and never put an element X that you see on top of another element Y in a layer that comes before the layer of Y. For example, if you see text on top of a photo, it should be in a layer that comes after the layer of the photo.
    Please start from the background and proceed gradually to what you see on top. For example, if you see text on top of an image or an illustration, you must first mention the image or illustration, and then the text. This is very important.
    Please do not mix text with other elements. If there is text, it should be separated from the other visual elements like photos, shapes, icons, etc.
    The format you must follow is a short description of the design, followed by the layers, in a list. Do not add any other element or hint, just the description and the layers. Use a single list, without sublists.

    The prompt is: \{image\_description\_content\}

    Please organise the elements in the photo into  \{num\_layers\} layers. It is mandatory to have only  \{num\_layers\}. If more information is needed  to describe the layer, add more sentences one after the other.
     
\end{tcolorbox}
\caption{}\label{prompt:i2l-2}
\end{figure*}

\begin{figure*}[t]
\centering
\begin{tcolorbox}[title=Prompt Expansion]
Your task is to create structured design descriptions based on provided problem statements. Follow these specific guidelines to ensure your responses meet the required format and criteria:

1. **Overall Description**: Start your design description with a brief overview of the overall design concept before listing the individual layers. This should provide context and purpose for the design.

2. **Layered Structure**: Present your answer in a clear, layered format. Each layer should begin with a dash (-) and represent a distinct element of the design. Ensure that the layers are organized logically and sequentially, with a maximum of 6 layers.

3. **Quotation Usage**: Enclose all actual text elements (e.g., titles, names, dates, URLs) in quotation marks. This includes any promotional text, headers, or important details that need to be highlighted.

4. **Text Implementation**: Ensure that all relevant text is included and properly quoted. This includes not only the main titles but also any additional information such as contact details, event specifics, or descriptions.

5. **Visual Design Elements**: Describe the visual aspects of the design clearly. This includes background colors, layout arrangements, and any graphic elements that are part of the design. Be specific about the colors, shapes, and positioning of elements.

6. **Clarity and Professionalism**: Your descriptions should be professional and cohesive. Ensure that all elements are clearly described and that the overall design concept is easy to understand.

7. **Avoid Headers and Nested Bullets**: Do not use headers or nested bullet points. The structure should be flat, with each layer clearly delineated by a dash.

8. **Example Format**: When providing your answer, follow this example format:
   A clean and professional promotional poster with a minimalist design approach.
   - a solid white background. \\
   - a bold title "TITLE" in large font. \\
   - additional text "DETAILS" in smaller font. \\
   - any relevant images or graphics described clearly. \\

9. **Design Evaluation Criteria**: Be aware of the following evaluation criteria to ensure your design descriptions are effective: \\
   - **Overall Description**: Start with a brief overview of the overall design concept before listing layers. \\
   - **Layer Count**: Aim for a maximum of 6 layers to maintain clarity and conciseness. \\
   - **Layer Format**: Ensure all layers begin with a dash and avoid any headers or nested formats. \\
   - **Quotation Usage**: Only actual text should be quoted, ensuring clarity and emphasis on important elements. \\
   - **Text Implementation**: Include all necessary text elements and ensure they are properly quoted. \\
   - **Simple Structure**: Maintain a flat structure without nested bullets for ease of reading. \\
   - **Professionalism**: Ensure the design is cohesive and intentional, reflecting a clear understanding of the task. \\
   - **Clarity**: All elements should be described clearly, allowing for easy interpretation of the design.

By adhering to these guidelines, you will create effective and professional design descriptions that meet the specified criteria. Always refer back to these instructions when tackling similar tasks in the future.
\end{tcolorbox}
\caption{}\label{prompt:pe}
\end{figure*}

\end{document}